\documentclass[10pt,journal,compsoc]{IEEEtran}
\usepackage[table,dvipsnames]{xcolor}
\usepackage[table]{xcolor}
\RequirePackage[dvipsnames]{xcolor}
\usepackage{amsmath,amsfonts}

\usepackage{array}
\usepackage[caption=false,font=normalsize,labelfont=sf,textfont=sf]{subfig}
\ifCLASSOPTIONcaptionsoff
 \usepackage[nomarkers]{endfloat}
\let\MYoriglatexcaption\caption
\renewcommand{\caption}[2][\relax]{\MYoriglatexcaption[#2]{#2}}
\fi
\usepackage{textcomp}
\usepackage{stfloats}
\usepackage{url}
\usepackage{verbatim}
\usepackage{graphicx}
\def\BibTeX{{\rm B\kern-.05em{\sc i\kern-.025em b}\kern-.08em
    T\kern-.1667em\lower.7ex\hbox{E}\kern-.125emX}}
\usepackage{balance}
\usepackage[caption=false,font=normalsize,labelfont=sf,textfont=sf]{subfig}
\usepackage{textcomp}
\usepackage{stfloats}
\usepackage{url}
\usepackage{verbatim}
\usepackage{graphicx}
\usepackage{soul}
\usepackage{tcolorbox}
\usepackage{algorithm}
\usepackage{longtable}
\usepackage{supertabular}
\ifCLASSOPTIONcompsoc
  \usepackage[nocompress]{cite}
\else
  \usepackage{cite}
\fi
\usepackage{color}
\usepackage[normalem]{ulem}  
\usepackage{xspace}  
\usepackage{booktabs}
\usepackage{multirow}
\usepackage{pifont}
\usepackage{arydshln}
\usepackage{colortbl}
\usepackage{hyperref}
\usepackage{makecell,rotating,multirow,diagbox}
\makeatletter
\DeclareRobustCommand\onedot{\futurelet\@let@token\@onedot}
\def\@onedot{\ifx\@let@token.\else.\null\fi\xspace}
\makeatother

\usepackage{multirow}  
\usepackage{textcomp}  
\usepackage{makecell} 
\usepackage{enumitem}
\usepackage{color}
 \usepackage{xcolor}
\usepackage{colortbl}
\usepackage{algorithm}
\usepackage{algpseudocode}
\usepackage{amsmath}
\usepackage{amssymb}
\usepackage{mathtools}
\usepackage{amsthm}
\usepackage{graphicx}
\usepackage{booktabs}
\usepackage{adjustbox}
\usepackage{subcaption} 
\usepackage{enumitem}
\usepackage{setspace}
\usepackage{pifont}
\usepackage{caption}
\usepackage{calligra}
\usepackage{tabularx} 
\usepackage{hyperref}

\newcommand{\bestSR}[1]{\textcolor{red}{#1}}
\newcommand{\secondSR}[1]{\textcolor{red}{\uline{\textcolor{black}{#1}}}}

\usepackage{array}
\newcolumntype{C}[1]{>{\centering\arraybackslash}p{#1}}
 
\definecolor{mygray}{gray}{.9}
\definecolor{ggray}{RGB}{127,127,127}
\definecolor{reda}{RGB}{192,0,0}
\definecolor{redb}{RGB}{217,148,143}
\definecolor{myyellow}{RGB}{255,240,166}
\definecolor{mygreen}{RGB}{197, 224, 180}
\definecolor{myblue}{RGB}{30,90,100}
\definecolor{mygray1}{RGB}{245,245,245}

\begin{document}

\title{See, Plan, Rewind: Progress-Aware Vision-Language-Action Models for Robust Robotic Manipulation}


\author{
  Tingjun Dai*\thanks{*Authors contribute equally. $^{\dag}$Project lead.},~Mingfei Han*$^{\dag}$,~Tingwen Du,~Zhiheng Liu,~Zihao Zhang,~Zhihui Li,\\~Salman Khan,~Jun Yu,~Xiaojun Chang,~\IEEEmembership{Senior~Member,~IEEE}
\IEEEcompsocitemizethanks{
\IEEEcompsocthanksitem Tingjun Dai, Mingfei Han, Tingwen Du, Zhihui Li and Xiaojun Chang are with School of Information Science and Technology, University of Science and Technology of China, Anhui, China. E-mail: \{hmf282@gmail.com, dutw2023@mail.ustc.edu.cn,\{lizhihuics,xjchang\}@ustc.edu.cn\}
\IEEEcompsocthanksitem Tingjun Dai is also with University of Technology Sydney, Ultimo, NSW, Australia. E-mail: \{tingjun.dai@student.uts.edu.au\}
\IEEEcompsocthanksitem Mingfei Han, Salman Khan are with Department of Computer Vision, Mohamed Bin Zayed University of Artificial Intelligence, Abu Dhabi, United Arab Emirates. E-mail: \{mingfei.han, salman.khan\}@mbzuai.ac.ae.
\IEEEcompsocthanksitem Zhiheng Liu is with The University of Hong Kong, Hong Kong, China.
E-mail: \{zhihengl0528@connect.hku.hk\}.
\IEEEcompsocthanksitem Zihao Zhang is with Institute of AI for Industry, Chinese Academy of Sciences, Jiangsu, China.
E-mail: \{zhangzihao@ict.ac.cn\}.
\IEEEcompsocthanksitem Jun Yu is with School of Intelligent Science and Engineering, Harbin Institute of Technology (Shenzhen), Guangdong, China.
E-mail: \{yujun@hit.edu.cn\}.

}}

\markboth{IEEE Transactions on Pattern Analysis and Machine Intelligence}%
{Dai \MakeLowercase{\textit{et al.}}: See, Plan, Rewind: Progress-Aware Vision-Language-Action Models for Robust Robotic Manipulation}


\IEEEtitleabstractindextext{%
    \begin{abstract}
Measurement of task progress through explicit, actionable milestones is critical for robust robotic manipulation. This progress awareness enables a model to ground its current task status, anticipate verifiable intermediate states, and detect and recover from failures when progress stalls. To embody this capability, we introduce \textbf{S}ee, \textbf{P}lan, \textbf{R}ewind (SPR), a progress-aware vision-language-action framework that dynamically grounds language instructions into a sequence of spatial subgoals. SPR operates through a continuous core cycle, Seeing the current state and upcoming milestone, Planning a trajectory towards the next 2D waypoint, and Rewinding to a recoverable state upon failure by monitoring progress against the expected sequence. This closed-loop approach enables robust error correction without requiring additional training data or auxiliary models. Extensive experiments demonstrate the framework's effectiveness, generalization and robustness: SPR outperforms the MolmoAct baseline by 5\% on the LIBERO benchmark. On the challenging LIBERO-Plus benchmark with unseen instructions and initial states, SPR achieves state-of-the-art robustness with the smallest performance drop, surpassing OpenVLA-OFT and UniVLA, demonstrating superior out-of-distribution robustness.
\end{abstract} 
}

\maketitle

\section{Introduction}
\label{sec:introduction}

Robotic manipulation requires a continual, closed-loop interaction with a dynamic 3D environment. 
While existing approaches~\cite{OLAF,YAY, thinkact, aha, robofac, come_robot, reflect, grape} have enabled basic task execution and flexible behaviors, 
robust performance demands an agent to not only perceive and act, but also maintain a grounded, quantitative awareness of its progress toward a goal. 
We formalize this capability as \textbf{\textit{progress awareness}}, the ability to measure task execution against a sequence of concrete and actionable milestones.

\begin{figure}[h!]
  \centering
  \includegraphics[width=0.98\linewidth]{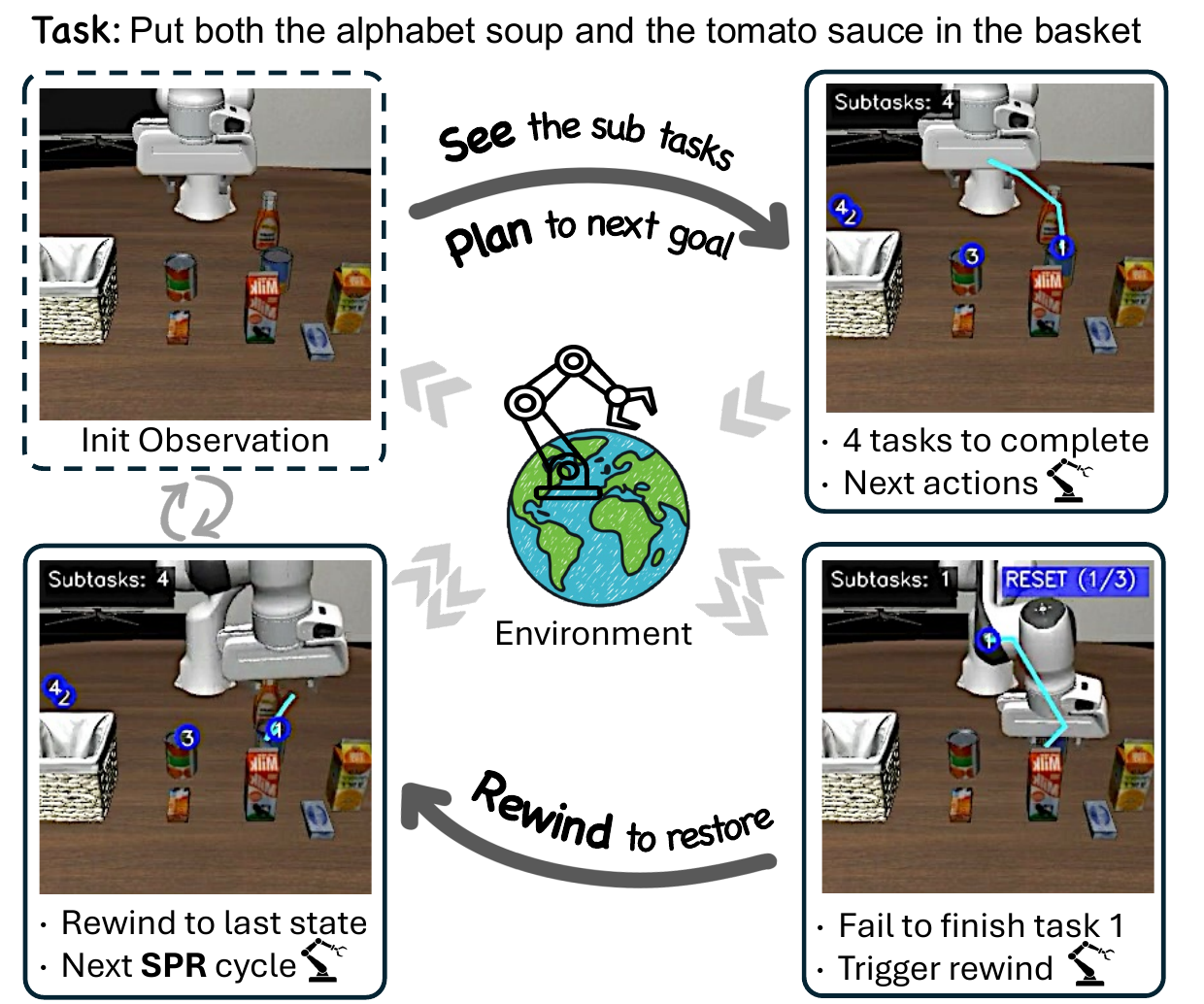}
   \caption{\textit{See-Plan-Rewind} framework's closed-loop execution workflow. Starting from the initial state (\textit{top-left}), the model performs See-Plan reasoning to generate actions (arrow to \textit{top-right}), which visualizes the subtask decomposition and trajectory planning. Under normal execution, the loop returns to top-left. When progress monitoring detects failures (\textit{bottom-right}), the Rewind mechanism activates (\textit{bottom-left}), returning the robot to the initial state before resuming the loop. This closed-loop design enables autonomous error recovery through explicit progress awareness.}
   \label{fig:1}
\end{figure}

Recent research recognized this need and explored progress monitoring capabilities. 
Several approaches \cite{ecot, molmoact, seqvla} have explored progress monitoring, such as ECOT \cite{ecot} which connects semantics to features via visual chain-of-thought, yet its progress signals remain abstract and lack spatial grounding.
In failure recovery, approaches include generating scalable failure-reasoning data \cite{failsafe, aha, robofac} and employing LLMs for error analysis and correction \cite{reflect, come_robot}. However, methods like FailSafe \cite{failsafe} rely on extensive additional failure data collection, which is costly, while REFLECT \cite{reflect} depends heavily on pre-defined LLM prompting, limiting adaptability in unseen scenarios.
While these approaches advance progress awareness, their signals often remain abstract linguistic constructs or binary flags, lacking explicit spatial grounding for robot action. Also, their recovery mechanisms typically depend on auxiliary models or substantial data collection. 
This gap underscores the need for a unified framework that provides robot-interpretable, spatially-grounded progress planning with an intrinsically embedded and data-efficient recovery mechanism.

To address these limitations, we introduce \textbf{See, Plan, Rewind (SPR)}, a framework that endows vision-language-action models with explicit completion awareness of task milestones through concrete spatial subgoals. Unlike abstract descriptions and plans lacking measurable progress benchmarks, SPR leverages gripper interactions from existing demonstrations to decompose tasks into sequences of grounded spatial waypoints. Each waypoint serves as a verifiable intermediate target that provides perceptual anchoring, simplifies trajectory planning, and enables unambiguous progress evaluation. Specifically, SPR first \textit{See}s to identify remaining subtasks, then \textit{Plan}s 2D trajectories to the next sub-goal bounded waypoint, and finally \textit{Rewind}s to restore in-distribution states upon detecting anomalies, without additional training data or auxiliary models.

To establish the explicit awareness, we propose an extensible pipeline that automatically decomposes tasks into spatially-grounded subgoals from demonstrations. For pick-and-place tasks, subtask boundaries are identified directly from gripper state transitions; for other manipulation types, we employ Gemini-3 to annotate subtask segments with boundary frames and semantic descriptions. For each segment, we extract 2D subgoal coordinates by leveraging DINOv3~\cite{dinov3} for gripper feature matching and SAM~\cite{sam} for precise segmentation. Finally, the model learns to associate observations with spatial subgoals, task status, and motion trajectories within structured behavioral sequences.
During execution, progress is monitored in a closed loop via a state recorder that continuously tracks predicted subtask counts and planned 2D trajectories. Sustained count increases indicate execution failures such as repeated failed grasps, while unchanged trajectories over extended timesteps signal progress stagnation where the robot remains trapped in unproductive states due to collisions, misalignment, or other environmental constraints. Both anomaly types trigger the Rewind mechanism to restore operational stability.


We evaluate SPR on both simulation and real-robot settings. On the LIBERO~\cite{libero} benchmark, SPR surpasses MolmoAct baseline by 3.8\%, and in the one-policy-for-all setup achieves an additional 1.2\% improvement, in contrast to methods like OpenVLA-OFT~\cite{openvla_oft} which regress in this challenging scenario. On the out-of-distribution LIBERO-Plus suite~\cite{libero_plus} with over 6800 test-time variants, SPR maintains minimal performance drop across variation types (18.8\% average v.s. 27.0\% for OpenVLA-OFT~\cite{openvla_oft} and 37.5\% for UniVLA~\cite{univla}), establishing new state-of-the-art cross-domain robustness. We further validate SPR on three real-robot tasks, including one basic pick-and-place task and two challenging scenarios involving long-horizon multi-object tidying and continuous-contact pushing, where the baseline fails entirely on both challenging tasks while SPR maintains consistent performance.

To summarize, we make the following contributions:\begin{itemize}
\item \textbf{Progress Awareness with Spatial Subtasks}: We establish a new paradigm for progress monitoring by decomposing tasks into a sequence of 2D spatial subgoals, replacing abstract plans with concrete, verifiable waypoints that enable fine-grained, robot-executable progress tracking without auxiliary models.

\item \textbf{Progress-Driven Error Recovery}: 
We formulate progress monitoring as an executable recovery policy that detects anomalies through progress tracking and restores the robot to in-distribution states.

\item \textbf{Effetiveness and OOD Robustness}: 
We demonstrate that SPR achieves superior performance and generalization, outperforming strong baselines on the LIBERO benchmark and setting a new state-of-the-art for out-of-distribution robustness on LIBERO-Plus suite with minimal performance degradation.
\end{itemize}

\section{Related Works} 
\noindent \textbf{Vision-Language-Action Models.}
Building on the capabilities of pretrained vision foundation models and large language models, vision-language models (VLMs) \cite{paligemma, prismatic, flamingo, yang2025walking, yu2025learning} have demonstrated strong performance in multimodal understanding. 
Motivated by these strengths, researchers have begun treating robot control as an additional output modality, fine-tuning VLMs on large-scale robot datasets \cite{oxe, bridgev2, droid, rt_1, spatialvla} to transfer their end-to-end prediction abilities to robotic tasks.  
Pioneering work RT-2 \cite{rt_2} co-fine-tunes on web-scale VQA and robot data, enhancing generalization and emergent capabilities.
OpenVLA \cite{openvla} provides an open-source counterpart that explores efficient fine-tuning via LoRA and quantization for deployment.
The $\pi_0$ model \cite{pi0} leverages high-quality data to improve performance and employs a flow-matching architecture to enhance real-time control capabilities. 
$\pi_{0.5}$ \cite{pi0.5} generalizes to long-horizon household tasks in unseen environments through co-training on multi-modal data and decomposes tasks into semantic subtasks for hierarchical decision-making. 
Errors in manipulation can accumulate over time, making them susceptible to propagation. Therefore, perceiving execution progress and enacting timely corrections is vital for robust robot policies.

\noindent \textbf{Progress Awareness and Execution Monitoring.}
Early progress monitoring relied on external human supervision, with systems like OLAF \cite{OLAF} using verbal guidance to synthesize recovery data and YAY \cite{YAY} updating policies through direct interventions, both facing scalability limitations.
Recent works have explored autonomous progress monitoring using VLMs\cite{ecot, molmoact, seqvla, gu2025safe} through various mechanisms. ECOT \cite{ecot} introduced visual chain-of-thought for subgoal decomposition and progress awareness. Other structured approaches include MolmoAct \cite{molmoact} generating mid-level spatial plans with sparse waypoints (though coarse sampling may limit precision) and SeqVLA \cite{seqvla} using a detection head for subtask completion awareness.
However, many VLM-driven approaches still depend on prompt-based heuristics and lack explicit grounding in dynamic physical task states, leading to unreliable progress awareness under visual ambiguity or occlusion.

\noindent \textbf{Failure Recovery and Robust Control.}
While these monitoring capabilities enhance task awareness, the occurrence of failures during execution remains inevitable, making autonomous recovery crucial for preventing error accumulation and ensuring overall task success.
Early failure recovery methods relied on external human knowledge\cite{YAY, OLAF, palempallihuman}, facing scalability limitations. With the emergence of VLMs, research shifted toward autonomous reasoning\cite{reflect, come_robot, chen2024automating, racer, star}, where REFLECT \cite{reflect} uses LLMs to interpret execution experiences and suggest corrections, while COME-robot \cite{come_robot} employs GPT-4V for adaptive replanning. Relying solely on VLMs’ reasoning often leads to unreliable recovery in novel environments, prompting methods like AHA \cite{aha}, RoboFAC \cite{robofac}, and FailSafe \cite{failsafe} to generate targeted failure-and-recovery data.
However, these data-driven methods require extensive failure data collection, which is costly. This motivates alternatives that leverage successful demonstrations to synthesize recovery behaviors, reducing dependency on failure-specific data.

\section{See-Plan-Rewind Framework}

\begin{figure*}
  \centering
  \includegraphics[width=1.0\textwidth]{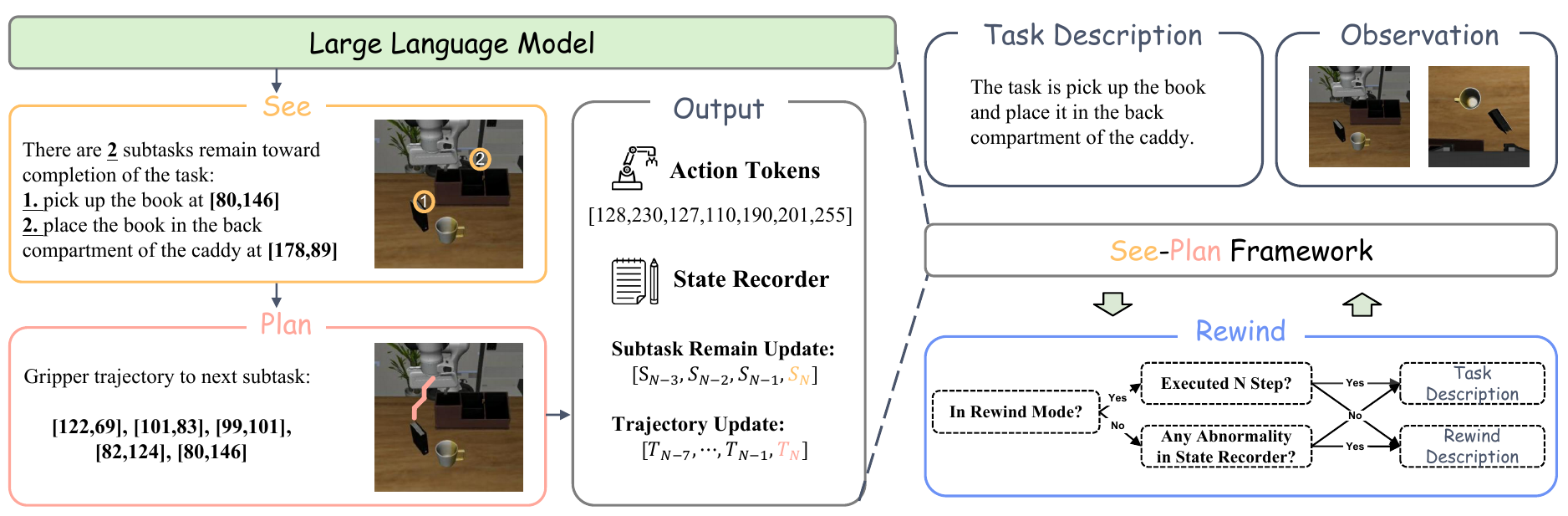}
  \caption{SPR framework overview. \textit{Left}: Upon receiving the task description and observation, the model performs See-Plan reasoning(Sec.~\ref{sec:subtask_planning}), which identifies remaining subtasks with 2D spatial coordinates (See) and plans a gripper trajectory to the next subtask waypoint (Plan), then outputs action tokens for execution. Each inference step also updates the state recorder, where $S_N$ denotes the predicted subtask count and $T_N$ the planned 2D trajectory at the current timestep. \textit{Right}: The Rewind mechanism (Sec.~\ref{sec:rewind}) examines the state recorder after each step: if no anomaly is detected, the original task description is retained; if sustained anomalies are identified, the task description is switched to a rewind instruction for $N$ steps before reverting to normal execution. Data generation in Sec.~\ref{sec:data_generation}.}
  \label{fig:2}
\end{figure*}

We introduce our See-Plan-Rewind framework, a progress-aware vision-language-action model that achieves robust manipulation through fine-grained spatial subtask planning and error recovery. As illustrated in Figure~\ref{fig:2}, our approach operates through a continuous cycle: the model first Sees the current state and subtask remain (Section~\ref{sec:subtask_planning}), Plans a trajectory towards next 2D waypoint, and Rewinds to a recoverable state when anomalies are detected (Section~\ref{sec:rewind}). The foundation of this framework lies in our comprehensive data generation pipeline (Section~\ref{sec:data_generation}), which automatically constructs supervision for subtask boundaries, spatial coordinates, and rewind trajectories without additional human annotation or auxiliary models.


\subsection{Fine-Grained Spatial Subtask Planning}
\label{sec:subtask_planning}

Grasping and manipulating objects is so routine for humans that we rarely notice the sophisticated progress awareness driving these actions. Our brain naturally decomposes manipulation goals into intermediate milestones, plans hand trajectories toward each subgoal, and continuously monitors execution progress—all without conscious thought. Inspired by this cognitive process, we design a fine-grained spatial subtask planning mechanism that endows VLA models with analogous progress-aware reasoning. By explicitly modeling the ``See-Plan'' cycle—identifying remaining subtasks and planning trajectories to concrete spatial waypoints—our framework enables robots to approach manipulation with the structured, milestone-driven strategy that humans employ effortlessly.

\subsubsection{Spatial Subtask See and Plan}

As illustrated in Figure~\ref{fig:2}, upon receiving the observation and task instruction, the model performs the See phase: outputting depth perception tokens (following MolmoAct, not shown in figure), predicting remaining subtask count, and sequentially generating semantic descriptions with 2D coordinates for each subtask, establishes comprehensive progress awareness. In the subsequent Plan phase, the model generates a 2D trajectory with up to 5 waypoints from the current gripper position to the next subtask goal. 
Unlike the baseline that plans directly toward the final goal, our next-subtask planning provides more robust execution guidance, particularly for long-horizon tasks where the final goal may be spatially irrelevant or even misleading before intermediate subtasks are completed, rendering end-to-end trajectory planning ineffective.

\subsubsection{Action Reasoning with Subtask Awareness}

Building upon MolmoAct's depth-conditioned reasoning, our framework extends the autoregressive generation process to incorporate explicit subtask planning. Given an RGB observation $I$ and language instruction $T$ (which includes the action CoT prompt), the model sequentially generates five token streams: (i) depth perception tokens $d$; (ii) remaining subtask count $n$; (iii) subtask specifications $s = (s_1, \ldots, s_n)$, where each $s_i = (\text{sem}_i, \text{coord}_i)$ consists of semantic description and 2D completion coordinates; (iv) trajectory tokens $\tau$ from current gripper position to the next subtask waypoint; (v) action tokens $a = (a_1, \ldots, a_D)$ with $D$ degrees of freedom. This factorization follows:
\begin{equation}
p(a, d, n, s, \tau | I, T) = p(d) \cdot p(n) \cdot p(s) \cdot p(\tau) \cdot p(a)
\label{eq:factorization}
\end{equation}
where each component conditions on the observation $I$, instruction $T$, and all previously generated tokens in the autoregressive sequence.

\subsection{See-Plan Data Curation}
\label{sec:data_generation}

Our See-Plan framework requires two types of supervision: (i) subtask boundaries and counts, (ii) spatial coordinates for subtask waypoints and trajectories. We design an automated pipeline to extract all supervision signals from existing demonstration data, eliminating the need for additional human annotation or auxiliary vision-language models during both training and inference.



\subsubsection{Subtasks Segmentation and Description.}
Our method segments demonstration trajectories into meaningful subtasks through task-adaptive boundary detection. For pick-and-place tasks, we identify subtask boundaries directly from gripper state transitions (open/close), which reliably mark subtask completion points. For tasks where gripper actuation alone is insufficient to delineate subtasks (\textit{e.g.}, pushing, closing cabinets), we employ Gemini-3 to generate subtask annotations, including start and end frame indices and semantic descriptions for each segment. For each frame, we compute the remaining number of subtasks, which serves as ground truth for training the model's progress monitoring. As a special case, since LIBERO~\cite{libero} tasks are predominantly pick-and-place, we apply the gripper-based segmentation and prompt DeepSeek-R1~\cite{deepseek} with the task instruction and detected subtask count to generate semantic descriptions for each segment.

\subsubsection{Gripper Trajectory Extraction.}
To obtain spatial coordinates for subtask waypoints and 2D trajectories, we combine DINOv3~\cite{dinov3} and SAM~\cite{sam} for robust gripper detection without task-specific training.
We maintain a reference image of the gripper endpoint and use DINOv3's patch-level features to identify the image region with highest similarity. This coarse localization is then refined by SAM, using the previously obtained position as point prompt for gripper segmentation. Followingly, we compute final coordinates by leveraging each method's strength: $x$ from SAM's bounding box center (precise horizontal boundaries) and $y$ from DINOv3 detection (accurate vertical endpoint localization). All 2D coordinates are discretized to [0, 255].
Finally, we smooth the trajectory by detecting and interpolating over outlier points with unusually large movements, and then applying a median filter with a small temporal window while preserving subtask boundaries.
Given gripper coordinates for all frames, we extract subtask waypoints as the gripper positions at detected boundary frames. We uniformly sample 1-5 gripper positions from the current frame to the next subtask completion frame, providing intermediate waypoints for trajectory planning.


\subsection{Error Recovery via Progress-Aware Rewind}
\label{sec:rewind}

While our subtask planning framework enables accurate progress monitoring, robust spatial planning alone cannot eliminate all failure modes. Inspired by human problem-solving strategies of ``stepping back'' when encountering obstacles, we propose a Rewind mechanism that disengages the model from erroneous states through a brief learned retraction, preventing the robot from persisting in failure modes with diminishing returns. Leveraging our framework's subtask counting and 2D trajectory grounding capabilities, we can detect potential execution anomalies in real time and trigger timely recovery—implemented entirely through joint training on constructed rewind data, without additional recovery demonstrations or auxiliary models.

\subsubsection{Rewind Data Construction}
To endow the model with rewind capability, we construct reverse trajectories by inverting successful forward demonstrations from the first subtask waypoint back to the robot's initial position. The construction involves: (i) temporal reversal of frame sequences, (ii) negation of action token values (inverting end-effector delta movements), and (iii) setting the task instruction to ``return to initial position.'' All other supervision signals—subtask boundaries, waypoint coordinates, and 2D trajectories—are automatically inherited from the data curation pipeline described in Section~\ref{sec:data_generation}.


\subsubsection{Progress-Based Anomaly Detection}
We maintain a state recorder, a first-in-first-out queue that continuously tracks the model's predicted subtask counts over the most recent 4 timesteps and the planned 2D trajectories over the most recent 8 timesteps. Anomalies are detected through two complementary criteria:

\noindent\textbf{Subtask Count Anomaly.} Under normal execution, the predicted subtask count either remains constant or decreases monotonically as subtasks complete. We detect anomalies by examining whether the subtask count increases across both the current and the immediately prior window in the queue, indicating sustained execution failure that causes the model to regress to earlier stages.

\noindent\textbf{Progress Stagnation.} If the planned 2D trajectories remain identical across all 8 recorded timesteps, we identify the robot as trapped in an out-of-distribution state caused by collisions, misalignment, or other unexpected environmental conditions. In such states, the model encounters configurations absent from its training data and fails to generate effective actions, resulting in repeated identical plans without meaningful progress toward the current subtask.

Both criteria require anomalies to persist over multiple timesteps, filtering transient prediction noise while reliably identifying sustained execution failures.

\subsubsection{Rewind Execution Strategy}

Through joint training on forward demonstrations and constructed rewind data, the model learns to retreat to its initial position upon receiving a ``return to initial position" instruction. During execution, if an anomaly is detected, we substitute the original task instruction with this rewind command for a fixed duration of $N$ timesteps. This allows the robot to backtrack toward its starting configuration. After $N$ steps, the system reverts to the original instruction and resumes the task from the new state. The duration $N$ is critical: a value too small fails to provide sufficient operational clearance, while a value too large risks the arm exiting the camera's view or assuming unrecoverable poses. We empirically set $N=3$ for optimal performance.

\section{Experiments}

\begin{table*}
  \caption{Out-of-distribution robustness evaluation on LIBERO-Plus benchmark. We report task success rates across five perturbation types, with subscripts indicating performance degradation relative to the original LIBERO test sets. \textbf{Bold} and \underline{underlined} values denote the best and second-best success rates, respectively. \bestSR{Red bold} and \secondSR{red underlined} subscripts highlight the smallest and second-smallest performance drops, indicating superior OOD robustness. Our method achieves the highest average success rate (71.8\%) with the smallest average degradation (18.8\%), demonstrating strong generalization to unseen perturbations.}
  \label{tab:libero_plus_results}
  \centering
  \small
  \setlength{\tabcolsep}{10pt}
  \resizebox{0.98\linewidth}{!}{%
  \begin{tabular}{@{}lcccccc@{}}
    \toprule
    \textbf{Method} & \multicolumn{6}{c}{\textbf{LIBERO-PLUS}} \\
    \cmidrule(lr){2-7}
    & \textbf{Background} & \textbf{Robot} & \textbf{Language} & \textbf{Layout} & \textbf{Light} & \textbf{Avg} \\
    \midrule
    OpenVLA\cite{openvla} & 25.3\%$_{\downarrow51.2\%}$ & 4.1\%$_{\downarrow72.4\%}$ & 26.8\%$_{\downarrow49.7\%}$ & 31.6\%$_{\downarrow44.9\%}$ & 4.4\%$_{\downarrow72.1\%}$ & 18.7\%$_{\downarrow57.8\%}$ \\
    OpenVLA-OFT\cite{openvla_oft} & 83.6\%$_{\downarrow14.0\%}$ & 30.6\%$_{\downarrow67.0\%}$ & \textbf{83.6\%}$_{\downarrow\secondSR{14.0\%}}$ & \textbf{73.2\%}$_{\downarrow24.4\%}$ & \textbf{91.6\%}$_{\downarrow\secondSR{6.0\%}}$ & \underline{70.6\%}$_{\downarrow27.0\%}$ \\
        OpenVLA-OFT\_w\cite{openvla_oft} & \textbf{92.5\%}$_{\downarrow\bestSR{\textbf{2.8\%}}}$ & 43.7\%$_{\downarrow51.6\%}$ & 73.2\%$_{\downarrow22.1\%}$ & \underline{72.3\%}$_{\downarrow23.0\%}$ & 68.2\%$_{\downarrow27.1\%}$ & 68.5\%$_{\downarrow26.8\%}$ \\
    $\pi_0$\cite{pi0} & 78.5\%$_{\downarrow15.7\%}$ & 6.6\%$_{\downarrow87.6\%}$ & 61.0\%$_{\downarrow33.2\%}$ & 70.4\%$_{\downarrow23.8\%}$ & 79.6\%$_{\downarrow14.6\%}$ & 56.6\%$_{\downarrow37.6\%}$ \\
    $\pi_0$-fast\cite{fast} & 67.7\%$_{\downarrow17.8\%}$ & 24.8\%$_{\downarrow60.7\%}$ & 63.3\%$_{\downarrow22.2\%}$ & 70.3\%$_{\downarrow\bestSR{\textbf{15.2\%}}}$ & 73.0\%$_{\downarrow12.5\%}$ & 58.4\%$_{\downarrow27.1\%}$ \\
    WorldVLA\cite{worldvla} & 14.5\%$_{\downarrow64.6\%}$ & 30.2\%$_{\downarrow48.9\%}$ & 44.2\%$_{\downarrow34.9\%}$ & 39.4\%$_{\downarrow39.7\%}$ & 29.4\%$_{\downarrow49.7\%}$ & 32.8\%$_{\downarrow46.3\%}$ \\
    Nora\cite{nora} & 50.5\%$_{\downarrow37.4\%}$ & 41.1\%$_{\downarrow46.8\%}$ & 67.0\%$_{\downarrow20.9\%}$ & 63.9\%$_{\downarrow24.0\%}$ & 31.0\%$_{\downarrow56.9\%}$ & 51.8\%$_{\downarrow36.1\%}$ \\
    UniVLA\cite{univla} & 80.0\%$_{\downarrow15.2\%}$ & \textbf{50.3\%}$_{\downarrow\secondSR{44.9\%}}$ & 71.8\%$_{\downarrow23.4\%}$ & 34.3\%$_{\downarrow60.9\%}$ & 59.1\%$_{\downarrow36.1\%}$ & 57.7\%$_{\downarrow37.5\%}$ \\
    \midrule
    Ours & \underline{86.0\%}$_{\downarrow\secondSR{4.6\%}}$ & \underline{47.7\%}$_{\downarrow\bestSR{\textbf{42.9\%}}}$ & \underline{78.5\%}$_{\downarrow\bestSR{\textbf{12.1\%}}}$ & 69.6\%$_{\downarrow\secondSR{21.0\%}}$ & \underline{85.0\%}$_{\downarrow\bestSR{\textbf{5.6\%}}}$ & \textbf{71.8\%}$_{\downarrow\bestSR{\textbf{18.8\%}}}$ \\
    \bottomrule
  \end{tabular}}
\end{table*}

\begin{table}
  \caption{Performance on LIBERO benchmark. We report results for two training configurations: separately-trained models (Ours) fine-tuned individually on each subset, and a jointly-trained model (Ours$^{*}$) trained on all four subsets. \textbf{Bold} and \underline{underlined} values indicate the best and second-best results respectively.}
  \label{tab:libero_results}
  \centering
  \small
  \setlength{\tabcolsep}{4pt} 
  \begin{tabular}{@{}lccccc@{}}
    \toprule
    & \multicolumn{5}{c@{}}{\textbf{LIBERO}} \\
    \cmidrule{2-6}
    \textbf{Method} & \textbf{Spatial} & \textbf{Object} & \textbf{Goal} & \textbf{Long} & \textbf{Avg} \\
    \midrule
    Diffusion Policy\cite{diffusion_policy} & 78.3\% & 92.5\% & 68.3\% & 50.5\% & 72.4\% \\
    Octo \cite{octo}& 78.9\% & 85.7\% & 84.6\% & 51.1\% & 75.1\% \\
    OpenVLA \cite{openvla}& 84.7\% & 88.4\% & 79.2\% & 53.7\% & 76.5\% \\
    GRAPE \cite{grape}& 88.5\% & 92.1\% & 83.1\% & 57.2\% & 80.2\% \\
    ThinkAct \cite{thinkact}& 88.3\% & 91.4\% & 87.1\% & 70.9\% & 84.4\% \\
    $\pi_0$-fast \cite{fast}& \textbf{96.4\%} & \textbf{96.8\%} & 88.6\% & 60.2\% & 85.5\% \\
    MolmoAct \cite{molmoact}& 87.0\% & \underline{95.4\%} & 87.6\% & 77.2\%& 86.8\% \\
    \midrule
    \textbf{Ours} & 92.4\% & 93.0\% & \textbf{94.2\%} & \underline{82.8\%} & \underline{90.6\%} \\
    \textbf{Ours}$^{*}$ & \underline{93.2\%} & \underline{95.4\%} & \underline{93.2\%} & \textbf{85.4\%} & \textbf{91.8\%} \\
    \bottomrule
  \end{tabular}
\end{table}

We evaluate SPR in four key dimensions: 
\textbf{1) Overall Performance:} How does SPR compare to state-of-the-art VLA models on standard simulation benchmarks and real-robot tasks?
\textbf{2) OOD Robustness:} How does SPR perform across five diverse perturbation types in LIBERO-Plus (novel backgrounds, robot initial state, language phrasings, object layouts, and lighting conditions)?
\textbf{3) Ablations:} What are the individual contributions of spatial subgoal planning and the rewind mechanism? Does extended interaction time enable recovery from more complex failures?
\textbf{4) Visualization:} How does SPR decompose tasks and prove error recovery effective?

\subsection{Implementation Details}
\label{sec:imple_details}


We initialize from MolmoAct's mid-trained checkpoint and apply LoRA fine-tuning (rank 32, alpha 16) with action chunking following their post-training protocol. We detail the training configurations for both simulation and real-robot experiments below.

\noindent\textbf{Training on Simulation (LIBERO).} We evaluate two training configurations across 32 NVIDIA A100-80G GPUs. We apply the same data augmentation strategy as MolmoAct, including random cropping, resizing, and color jittering to improve robustness. Table~\ref{tab:training_config} summarizes the complete training configuration.

The \textbf{separately-trained models} fine-tune both vision encoder and language model on each LIBERO subset individually with batch size 128 and learning rate 5$\times$10$^{-4}$, training for 40K-80K steps until optimal validation performance is reached.

The \textbf{jointly-trained model} freezes the vision encoder and fine-tunes only the language model on all four subsets combined with a two-stage training process. First, we train on the original MolmoAct dataset for 20K steps with learning rate 1.5$\times$10$^{-4}$ to maintain broad manipulation capabilities. Then, we continue training on our SPR-annotated LIBERO data for 10K steps with learning rate 1.5$\times$10$^{-3}$ to learn progress-aware planning. During both stages, we sample data from the four subsets with a ratio of 5:5:4:8 (Spatial:Object:Goal:Long), proportional to their individual training requirements. All configurations use batch size 1024.

\noindent\textbf{Training with Real-Robot Tasks.} We collect 100 demonstration trajectories for \textit{Pick up the Object} and 200 trajectories each for \textit{Tidy up the Table} and \textit{Push-T}. Each task is trained separately on 4 NVIDIA A100-80G GPUs with batch size 64, learning rate 5$\times$10$^{-4}$, and 5,000 training steps. We freeze the vision encoder and fine-tune only the language model. The action chunk size is set to 4 during training, but only the first 2 actions are executed at each inference step to enable more responsive closed-loop control. All other hyperparameters remain consistent with the simulation setup. Table~\ref{tab:training_config_real} summarizes the real-robot training configuration.

\noindent\textbf{Inference.}
SPR maintains identical per-step inference cost as MolmoAct, sharing the same 7B architecture. The additional subtask planning outputs add minimal tokens compared to other predictions, resulting in negligible computational overhead. Furthermore, our error recovery mechanism operates through simple logical comparisons of subtask counts and trajectory records in the state recorder, requiring no additional model inference and thus having zero impact on execution speed. On 4$\times$RTX 4090 GPUs, our model achieves 2.08 Hz inference speed, closely matching the baseline while enabling robust error recovery.

\subsection{Environment Setup}
\label{sec:env_setups}

\begin{figure*}[t]
  \centering
  \includegraphics[width=0.9\textwidth]{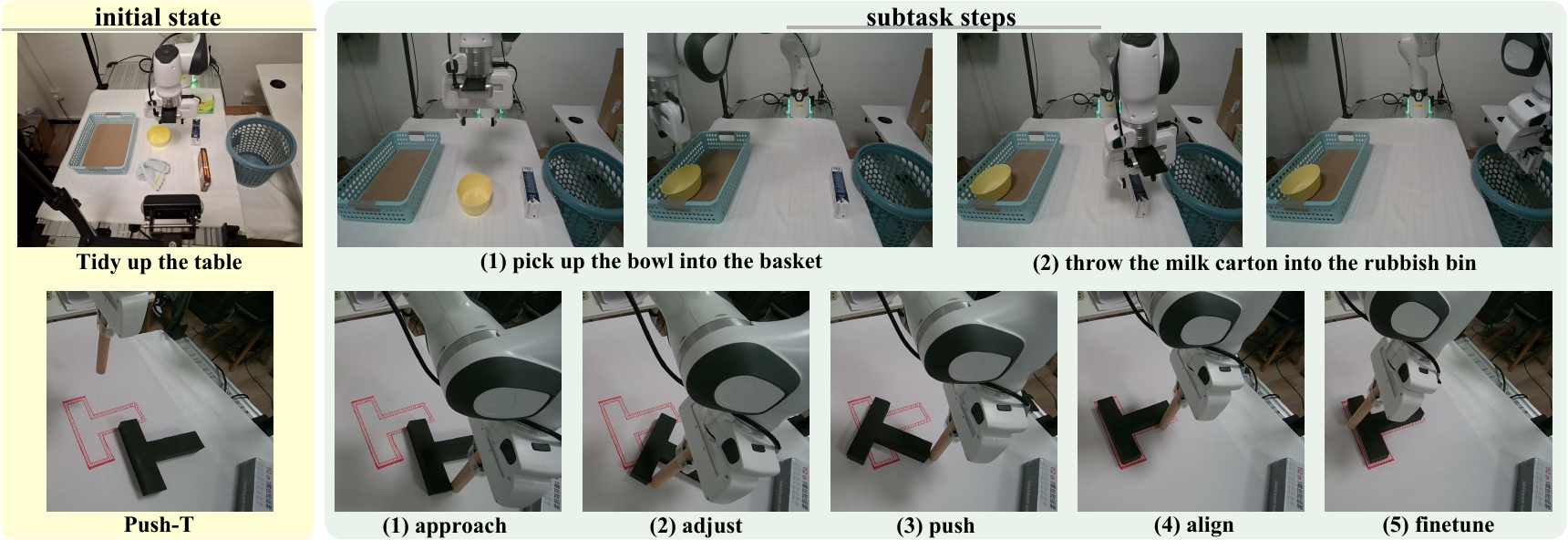}
  \caption{Real-robot task setup and subtask decomposition for \textit{Tidy up the Table} and \textit{Push-T}. Each task shows the initial scene configuration alongside the model's subtask decomposition, demonstrating that SPR produces structured and interpretable subtask plans for both pick-and-place and continuous-contact manipulation tasks.}
  \label{fig:real_visualization}
\end{figure*}




\begin{table}[t]
\caption{Training configuration for real-robot tasks. We collect 100 demonstrations for Pick up the Object and 200 each for Tidy up the Table and Push-T. All three tasks are trained separately with the vision encoder frozen. The action chunk size is 4 during training, but only the first 2 actions are executed at each inference step to enable more responsive closed-loop control.}
\label{tab:training_config_real}
\centering
\footnotesize
\renewcommand{\arraystretch}{1}
\begin{tabular}{@{}p{2.2cm}C{1.5cm}C{1.5cm}C{1.5cm}@{}}
\toprule
& \multicolumn{3}{c}{\textbf{Real-Robot Tasks}} \\
\cmidrule{2-4}
\textbf{Parameter} & \textbf{Pick up} & \textbf{Tidy up} & \textbf{Push-T} \\
\midrule
Demonstrations & 100 & 200 & 200 \\
Steps & 5K & 5K & 5K \\
Global Batch Size & 64 & 64 & 64 \\
Vision Encoder & \multicolumn{3}{c}{Frozen} \\
GPUs (A100s) & \multicolumn{3}{c}{4} \\
Input Images & \multicolumn{3}{c}{1 Third-person + 1 Wrist-mounted} \\
Image Size & \multicolumn{3}{c}{$256 \times 256$ px} \\
DoF & \multicolumn{3}{c}{7 (3 Translations + 3 Rotations + 1 Gripper State)} \\
Obs. History & \multicolumn{3}{c}{No (Single-step Inputs)} \\
Proprioception & \multicolumn{3}{c}{No} \\
Action Chunk & \multicolumn{3}{c}{4 Steps (Predict but Only Execute 2)} \\
\bottomrule
\end{tabular}
\end{table}

\begin{table}
\caption{Training configuration for LIBERO task suites. The first four columns show separately-trained models fine-tuned on individual subsets with both vision encoder and language model trainable. The last column shows the jointly-trained model with a two-stage process: first training on original MolmoAct data for 20K steps, then on our SPR-annotated data for 10K steps with data sampled at ratio 5:5:4:8 (Spatial:Object:Goal:Long).}
\label{tab:training_config}
\centering
\footnotesize
\renewcommand{\arraystretch}{1.}
\begin{tabular}{@{}p{2.2cm}C{0.8cm}C{0.8cm}C{0.8cm}C{0.8cm}C{1.1cm}@{}}

\toprule
& \multicolumn{5}{c}{\textbf{LIBERO Task Suite}} \\
\cmidrule(lr){2-6}
\textbf{Parameter} & \textbf{Spat.} & \textbf{Obj.} & \textbf{Goal} & \textbf{Long} & \textbf{All} \\
\midrule
Steps             & 50K  & 50K  & 40K  & 80K  & 20K+10K \\
Global Batch Size & 128  & 128  & 128  & 128  & 1024    \\
Vision Frozen & No   & No   & No   & No   & Yes     \\
GPUs (A100s)      & \multicolumn{5}{c}{32} \\
Input Images      & \multicolumn{5}{c}{1 Third-person + 1 Wrist-Mounted} \\
Image Size        & \multicolumn{5}{c}{$256 \times 256$ px} \\
DoF               & \multicolumn{5}{c}{7 (3 Trans.\ + 3 Rot.\ + 1 Grip.)} \\
Obs.\ History     & \multicolumn{5}{c}{No (Single-step Inputs)} \\
Proprioception& \multicolumn{5}{c}{No} \\
Action Chunk & \multicolumn{5}{c}{8 Steps (Predict and Execute 8)} \\
\bottomrule
\end{tabular}
\end{table}

\noindent \textbf{Simulation Benchmarks.} We evaluate our method on two robotic manipulation benchmarks:
\begin{itemize}[leftmargin=*,noitemsep,topsep=0pt]
    \item \textbf{LIBERO}~\cite{libero}: A widely-used benchmark suite for language-instructed manipulation in diverse kitchen scenarios. We report results on its four distinct subtask categories: \textit{Long} (complex multi-step tasks), \textit{Goal} (goal-conditioned tasks), \textit{Object} (object-centric manipulation), and \textit{Spatial} (tasks requiring spatial reasoning).
    \item \textbf{LIBERO-Plus}~\cite{libero_plus}: A challenging out-of-distribution benchmark that creates over 10,000 task variants across the four LIBERO test sets through seven types of perturbations. We evaluate on five subsets: \textit{Background} (unseen background texture), \textit{Robot} (unseen robot initial state), \textit{Language} (unseen instruction phrasing), \textit{Layout} (unseen object layout), and \textit{Light} (unseen scene lighting). Our model is trained exclusively on the original LIBERO training data, demonstrating strong generalization to these diverse unseen variations.
\end{itemize}

\noindent \textbf{Real-Robot Tasks.} We further validate SPR on three real-robot tasks, including one basic pick-and-place task and two challenging scenarios involving long-horizon multi-object tidying and continuous-contact pushing without grasping:
\begin{itemize}[leftmargin=*,noitemsep,topsep=0pt]
    \item \textbf{Pick up the Object}: Given a table with four objects (eggplant, biscuit box, toy, bowl), the robot must identify the correct object specified by the language instruction. This basic single-object task validates the model's fundamental language grounding and manipulation capability.
    \item \textbf{Tidy up the Table}: As shown in Figure~\ref{fig:real_visualization}, the workspace contains 1-4 target objects (towel, biscuit box, bowl, milk carton) along with distractors such as a tissue box. The robot must organize 1--4 objects (towel, biscuit box, bowl, milk carton) on the table into their designated receptacles—bowl and towel into a basket, biscuit box and milk carton into a rubbish bin—while ignoring distractors such as a tissue box. A trial is successful only when all objects are placed in their correct locations. This long-horizon setup with variable object counts serves to validate SPR's ability to plan and execute extended subtask sequences.
    \item \textbf{Push-T}: As shown in Figure~\ref{fig:real_visualization}, a T-shaped block and a target zone are placed on the table. 
    The robot uses a cylindrical end-effector to push a T-shaped block into a target zone, with success requiring full alignment. The figure illustrates how SPR decomposes this continuous-contact task into five subtask steps: approach, adjust, push, align, and fine-tune.
    This continuous contact manipulation task validates SPR's generalization beyond simple pick-and-place tasks.
\end{itemize}


\noindent \textbf{Evaluation Metric.} We evaluate task success rate, defined as the proportion of successful task completions. For LIBERO, we test 50 episodes per task across 10 tasks in each subset. For LIBERO-Plus, we test one episode per task across over 300 tasks for each of five perturbation types. For real-robot tasks, we conduct 10 trials per task configuration.



\begin{figure*}[t]
  \centering
  \includegraphics[width=1.\linewidth]{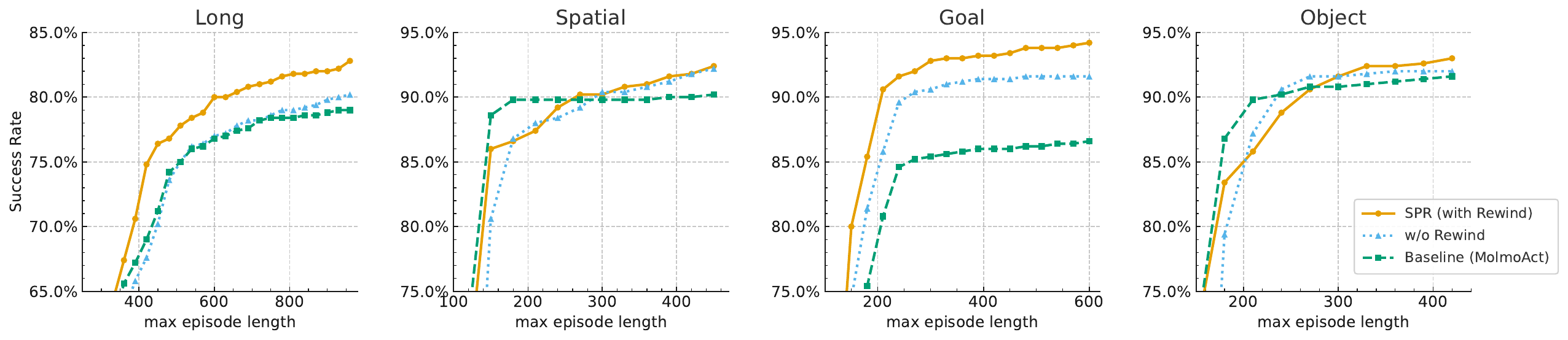}
   \caption{Task success rate vs. maximum episode length across LIBERO subsets. Progress-aware models continue improving after the baseline plateaus, demonstrating the ability to leverage extended horizons for complex error recovery.}
   \label{fig:4.3.1}
\end{figure*}

\begin{figure}[t]
  \centering
  \includegraphics[width=0.95\linewidth]{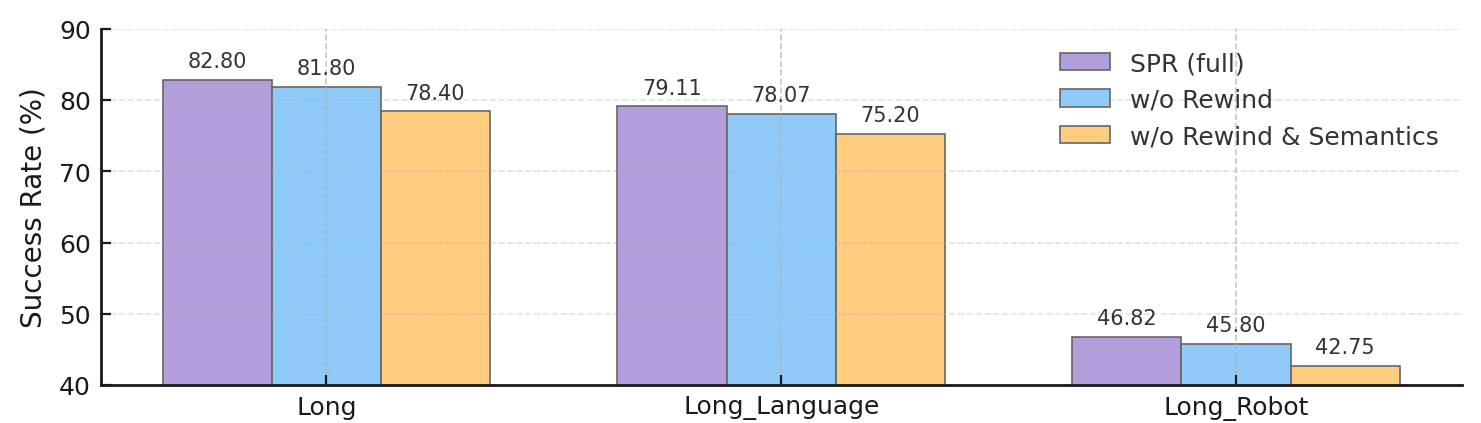}
   \caption{Component ablation on LIBERO-Long and LIBERO-Plus variants. Performance progressively improves from w/o Rewind \& Semantics (spatial coordinates only) to w/o Rewind (spatial + semantic) to Ours (full SPR with Rewind), validating the contribution of each component.}
   \label{fig:4.3.2}
\end{figure}

\subsection{Simulation Results}

\noindent\textbf{Effectiveness Evaluation on LIBERO.}
Table~\ref{tab:libero_results} presents SPR results on LIBERO. We evaluate two configurations: separately-trained models per subset (90.6\%, +3.8\% over MolmoAct) and jointly-trained on all subsets (91.8\%, +5.0\%). Both show strong gains on the challenging Long benchmark (+5.6\% and +8.2\% respectively), demonstrating that SPR effectively tackles complex multi-step manipulation. Moreover, the performance gain from joint training—rather than degradation—indicates that SPR learns generalizable progress-aware reasoning rather than overfitting to specific task distributions.

\noindent\textbf{Robustness Evaluation on LIBERO-Plus.}
Table~\ref{tab:libero_plus_results} presents results on the LIBERO-Plus benchmark under various distribution shifts. Notably, all state-of-the-art models exhibit substantial performance degradation on LIBERO-Plus compared to the original LIBERO test sets, underscoring the benchmark's difficulty. Among all evaluated methods, SPR demonstrates the smallest overall performance drop, highlighting its exceptional out-of-distribution robustness—a direct consequence of its fine-grained, geometry-grounded planning and built-in failure recovery mechanism. 
Particularly noteworthy, SPR achieves the most competitive performance degradation on Language (-12.1\%), Light (-5.6\%), and Robot (-42.9\%) perturbations. These results validate SPR's superior capability in handling semantic ambiguity and novel robot initial configurations. The strong performance under Language shifts demonstrates the effectiveness of our semantics-based progress awareness, while the robustness to Robot configuration changes highlights the value of our Rewind mechanism in adjusting gripper poses to recoverable states. Together, these capabilities enable SPR to maintain robust performance across diverse OOD scenarios, establishing superior zero-shot adaptation compared to existing approaches.

\begin{figure}[ht]
  \centering
  \includegraphics[width=1.\linewidth]{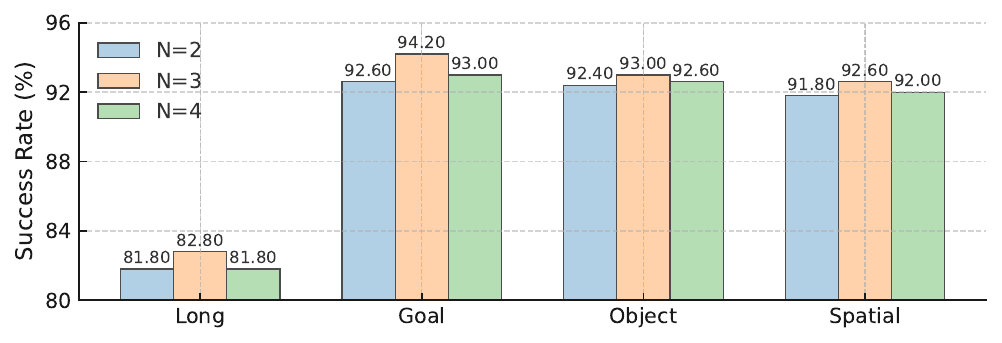}
   \caption{Effect of rewind step count $N$ on LIBERO. $N$=3 achieves optimal performance across all subsets.}
   \label{fig:4.3.3}
\end{figure}

\begin{figure}[ht]
  \centering
  \includegraphics[width=0.95\linewidth]{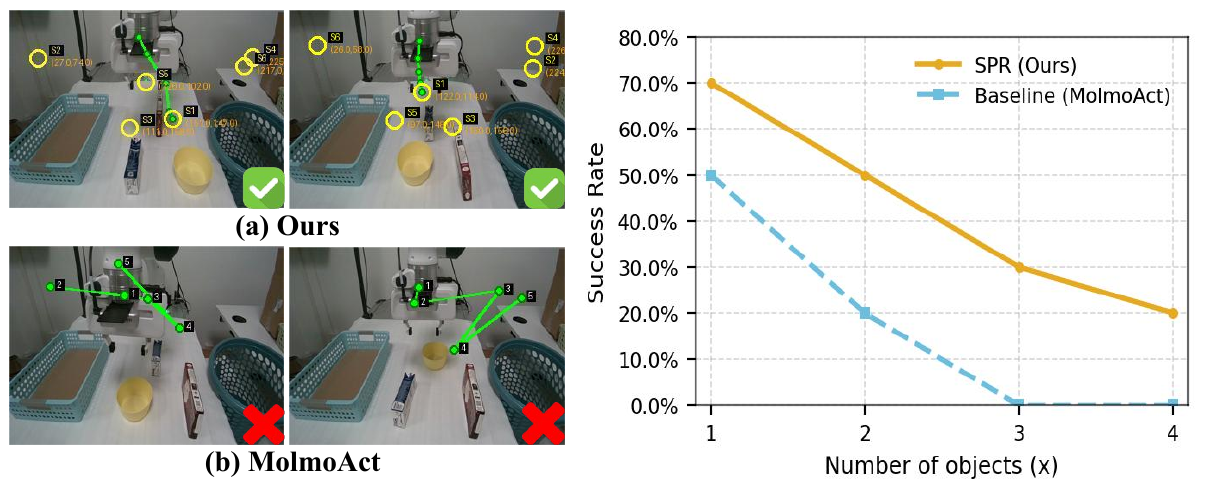}
   \caption{Scaling analysis on real-robot \textit{Tidy up the Table} with 1--4 objects. The planning visualization (left) compares SPR and MolmoAct on a 3-object trial, where yellow circles denote subtask waypoints and green lines indicate planned trajectories. The success rate curve (right) shows that SPR degrades gracefully as object count increases, while MolmoAct's coarse-grained planning collapses beyond 2 objects.}
   \label{fig:tidy_ablation}
\end{figure}

\subsection{Real-Robot Results}
Table~\ref{tab:real_robot} summarizes results across three tasks. Additional setup details and visualizations for all real-robot tasks are provided in the supplementary material.

\noindent\textbf{Pick up the Object.}
On this basic single-object task, SPR achieves 70\% success rate compared to 50\% for MolmoAct. Even for a single-step task, SPR's explicit spatial grounding of the target object as a subtask waypoint enhances the model's comprehension of the task goal and provides more precise trajectory planning toward the grasping target.

\noindent\textbf{Tidy up the Table.}
In the 3-object setting, SPR achieves 30\% success rate while MolmoAct fails entirely. SPR's subtask planning decomposes this complex long-horizon task into clearly ordered steps, guiding the model to complete them incrementally. In contrast, the baseline's coarse planning from the start to the final goal produces 2D waypoints that become noise rather than guidance as task complexity grows, ultimately failing to provide meaningful execution signals. A detailed analysis of how performance scales with object count is presented in Section~\ref{sec:ablation}.

\noindent\textbf{Push-T.}
SPR achieves a 40\% success rate while MolmoAct scores 0\%. This task involves continuous-contact manipulation where gripper state transitions cannot delineate subtask boundaries, yet SPR successfully decomposes it into five sequential phases (approach, adjust, push, align, and fine-tune), progressively guiding the T-block toward the target zone. This confirms that SPR is not confined to pick-and-place tasks, but can perceive and track progress across diverse manipulation types through its extensible subtask planning framework.

\subsection{Ablation Study}
\label{sec:ablation}

\noindent\textbf{Effect of Spatial Subgoals and Semantics.}
We first investigate whether subtask-level semantic planning can improve model performance—specifically, the model with only See and Plan capabilities but without Rewind. As shown in Table~\ref{tab:ablation_table}, comparing with the baseline MolmoAct model across all LIBERO test sets, the model with only See and Plan (w/o Rewind) achieves an overall 4.0\% performance improvement over the baseline, strongly demonstrating the effectiveness of our progress-aware approach.

To assess the individual importance of spatial coordinates and semantic descriptions, we evaluate w/o Rewind \& Semantics, which retains 2D coordinates but removes semantic generation. Figure~\ref{fig:4.3.2} reveals that both components are essential. While w/o Rewind \& Semantics outperforms the baseline on LIBERO-Long by 6\%—demonstrating spatial coordinates' contribution—it underperforms the w/o Rewind model by 3.4\%. The gap also exists on LIBERO-Plus, where removing semantics significantly degrades robustness. These results demonstrate that spatial coordinates and semantic descriptions play complementary roles: coordinates provide precise spatial grounding, while semantics enable higher-level task understanding. Both are crucial for effective progress-aware reasoning.

\noindent\textbf{Effect of Rewind Mechanism.}
We ablate the rewind mechanism to evaluate its contribution. As shown in Table~\ref{tab:ablation_table}, our full model achieves higher or comparable success rates across all LIBERO subsets compared to the model without rewind, with an overall performance gain of 1\%.
To further test its robustness, we evaluate on the more challenging Long suite of the LIBERO-Plus~\cite{libero_plus} benchmark, which introduces test-time variance. Results in Figure~\ref{fig:4.3.2} show that our model with rewind maintains superior performance on both its Language and Robot variants. These findings confirm that the rewind mechanism consistently enhances performance and provides critical robustness in complex, out-of-distribution scenarios.

\begin{table}[t!]
  \caption{Performance on real-robot tasks. We report success rates for our method and the MolmoAct baseline across three tasks: Pick up the Object, Tidy up the Table (evaluated in the 3-object setting), and Push-T. Additional task configurations and visualizations are provided in the supplementary material. \textbf{Bold} values indicate the best results.}
  \label{tab:real_robot}
  \centering
  \small
  \setlength{\tabcolsep}{10pt}
  \begin{tabular}{@{}lccc@{}}
    \toprule
    & \multicolumn{3}{c@{}}{\textbf{Real-Robot Tasks}} \\
    \cmidrule{2-4}
    \textbf{Method} & \textbf{Pick up} & \textbf{Tidy up} & \textbf{Push-T}  \\
    \midrule
    MolmoAct \cite{molmoact} & 50\% & 0\% & 0\%  \\
    \textbf{Ours} & \textbf{70\%} & \textbf{30\%} & \textbf{40\%}  \\
    \bottomrule
  \end{tabular}
\end{table}

\begin{table}[t!]
  \caption{Ablation study on LIBERO benchmark. Both See-Plan capabilities (w/o Rewind) and the complete SPR framework (Ours) outperform the MolmoAct baseline across all subsets.}
  \label{tab:ablation_table}
  \centering
  \small
  \setlength{\tabcolsep}{5pt}
  \begin{tabular}{@{}lccccc@{}}
    \toprule
    \textbf{Method} & \textbf{Spatial} & \textbf{Object} & \textbf{Goal} & \textbf{Long} & \textbf{Avg} \\
    \midrule
    MolmoAct$^{\dagger}$ & 89.4\% & 92.4\% & 88.2\% & 72.4\% & 85.6\% \\
    w/o Rewind & \textbf{92.6\%} & 91.8\% & 92.2\% & 81.8\% & 89.6\% \\
    \textbf{Ours} & 92.4\% & \textbf{93.0\%} & \textbf{94.2\%} & \textbf{82.8\%} & \textbf{90.6\%} \\
    \bottomrule
  \end{tabular}
  \vspace{2pt}
  
  {\footnotesize $^{\dagger}$Results reproduced using our training setup; differs from original~\cite{molmoact}.}
\end{table}

\noindent\textbf{Impact of Inference Episode Length.}
Our work endows the model with progress-aware capability, granting it enhanced retry ability after failed task executions. This raises a natural question: if we allow the model more generous step limits, providing additional opportunities to complete tasks, will it achieve higher success rates? To investigate this, we evaluate three model variants on all four LIBERO subsets with a maximum episode length of 980 steps: the baseline MolmoAct model, the w/o Rewind model, and our best model. As shown in the figure~\ref{fig:4.3.1}, across all LIBERO subsets, both our best model and the w/o Rewind model demonstrate superior ability to leverage extended episode lengths compared to the baseline, consistently achieving higher task success rates. Notably, after the baseline model's success rate plateaus, our models continue to complete additional tasks. This validates our model's progress-aware capability, enabling recovery from more complex error scenarios.

\begin{figure*}[t]
  \centering
  \includegraphics[width=1.\textwidth]{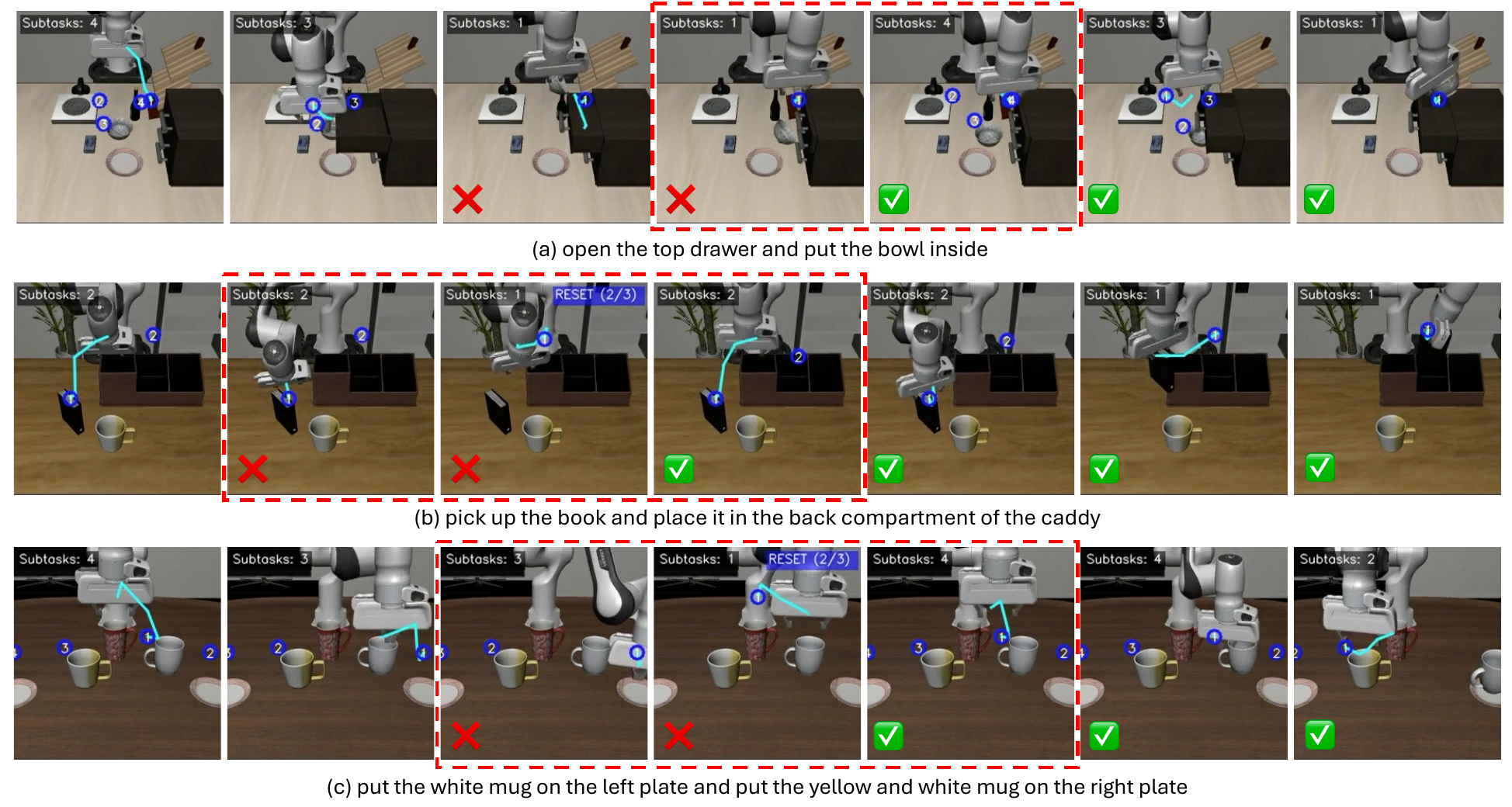}
  \caption{SPR's error recovery across diverse failure scenarios (red dashed boxes: error states). (a) \textit{Dynamic replanning}: Recovers from object relocation and environment state changes by updating spatial subtasks. (b) \textit{Robustness to suboptimal initial starts}: Mitigates challenging initial configurations by rewinding to regain spatial freedom and replanning the approach. (c) \textit{Execution failure recovery}: Detects OOD states from failed grasps and resets to familiar configurations for retry.}
  \label{fig:robust_drawer}
\end{figure*}

\begin{figure}[t]
  \centering
  \includegraphics[width=0.48\textwidth]{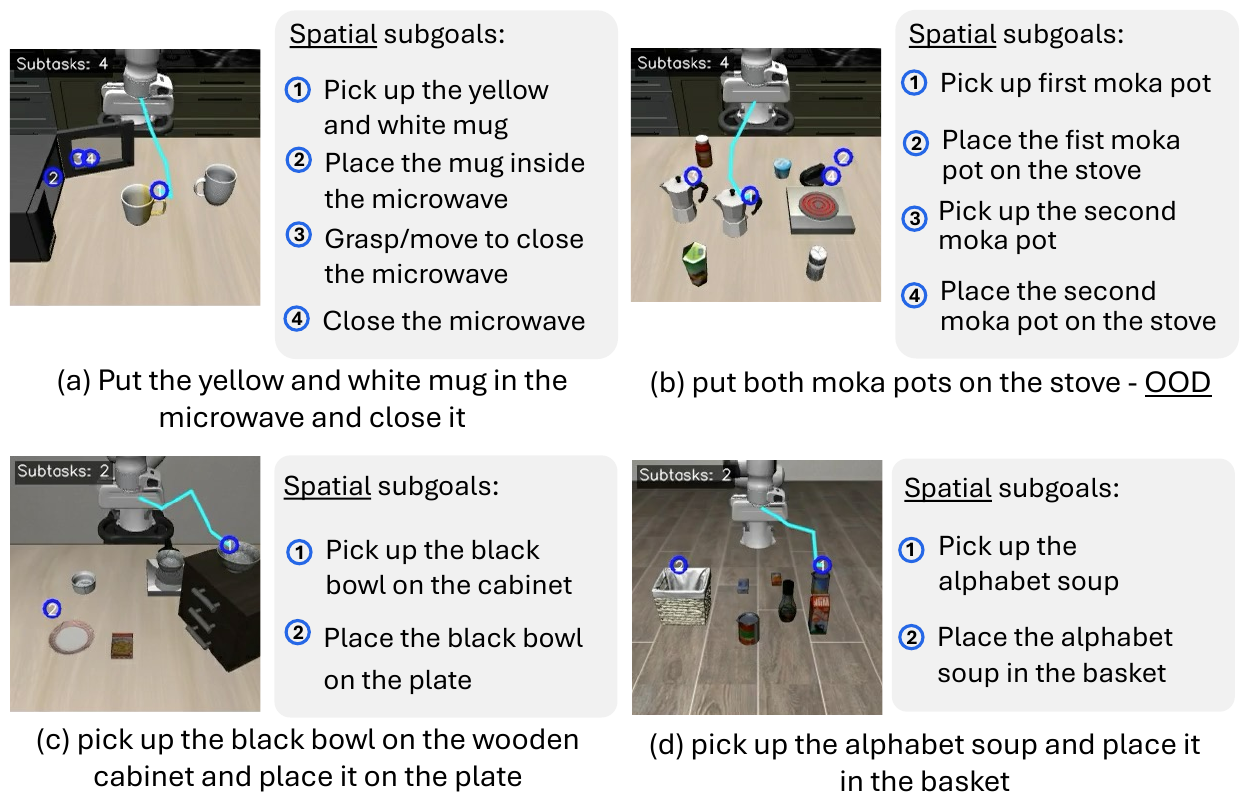}
  \caption{Fine-grained spatial subtask decomposition across diverse manipulation tasks. (a) Complex sequential task with multiple interaction types. (b) Robust decomposition under OOD layout configurations. (c) Spatial relationship understanding with relative positioning. (d) Basic pick-and-place with clear semantic grounding. Each subtask pairs semantic descriptions with 2D spatial coordinates, enabling interpretable progress monitoring.}
  \label{fig:subtask_semantics}
\end{figure}

\begin{figure*}[h!]
  \centering
  \includegraphics[width=0.9\textwidth]{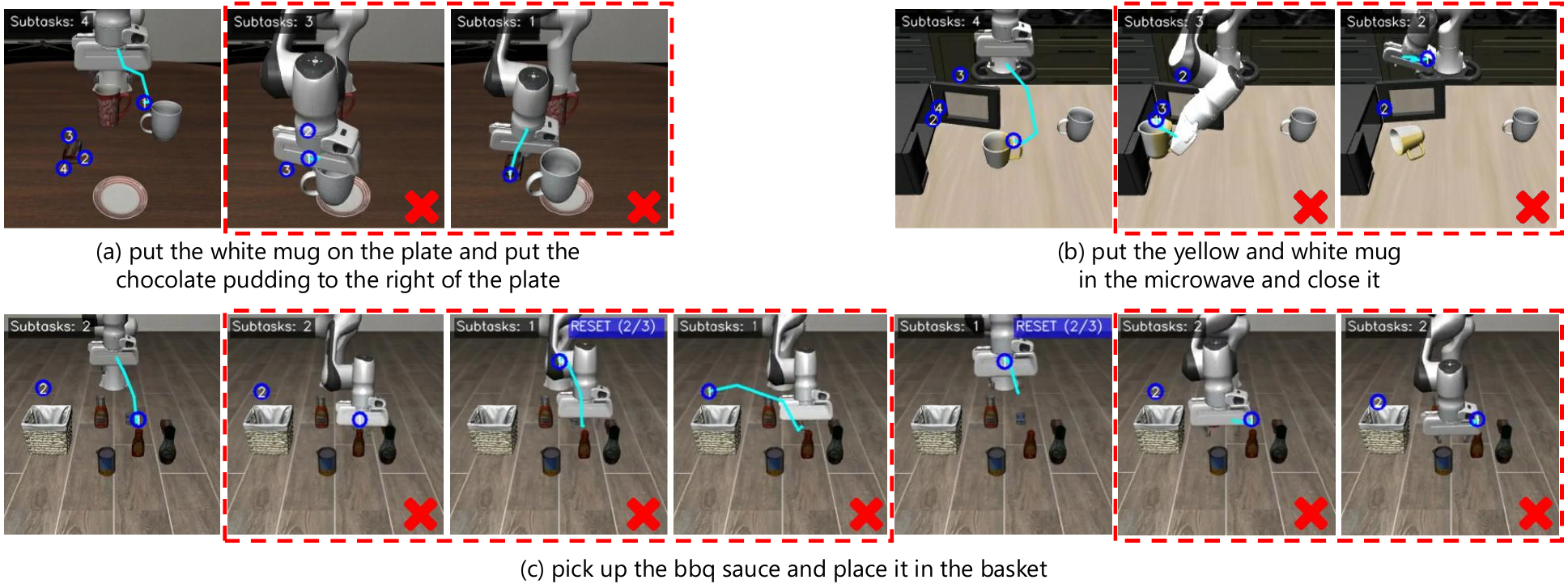}
  \caption{Representative failure cases: (a) discrete action tokens limit precision for careful placement tasks, (b) rewind mechanism fails when robot is physically stuck, and (c) model fails to complete task despite successful rewind due to persistent spatial misalignment and action-planning inconsistency.}
  \label{fig:failure_case}
\end{figure*}

Furthermore, we observe that our best model exhibits faster task completion efficiency compared to the w/o Rewind model. This demonstrates that the Rewind mechanism effectively reduces time required to re-execute failed subtasks, validating the theoretical correctness of elevating the arm to provide greater operational and observational space for improved error recovery. Especially on the two more complex subsets—LIBERO-LONG and LIBERO-Goal—our best model achieves significantly faster task completion efficiency compared to both other variants.

\noindent\textbf{Effect of Rewind Step Count N.}
When the Rewind mechanism triggers, the model executes $N$ consecutive ``return to initial position" instructions. We investigate the optimal value by testing $N \in \{2, 3, 4\}$ across all LIBERO test sets (Figure~\ref{fig:4.3.3}). Results show $N=3$ consistently achieves the best performance. Too few steps ($N<3$) fail to provide sufficient operational space for error recovery, while too many steps ($N>3$) cause the arm to drift excessively far from the task region. Moreover, we observe that continuous rewind instructions beyond three steps cause pose distortion, with the arm eventually moving outside the camera's field of view. Thus, $N=3$ balances effective error recovery with pose stability.

\noindent\textbf{Scaling to Long-Horizon Real-Robot Tasks.}
To further investigate SPR's advantage in extended real-world task sequences, we evaluate \textit{Tidy up the Table} with object counts from 1 to 4. Shown in Figure~\ref{fig:tidy_ablation}, as the number of objects increases, both SPR and MolmoAct experience performance degradation, but SPR degrades more gracefully. When object count reaches 3 and above, MolmoAct fails entirely while SPR continues to achieve successful completions. This widening gap confirms that subtask-level progress planning becomes increasingly critical as task complexity grows, and that SPR's structured decomposition scales effectively to challenging long-horizon scenarios.

\subsection{Visualization}

We present representative examples demonstrating SPR's fine-grained spatial planning and robust error recovery across diverse scenarios.

\noindent\textbf{Fine-Grained Spatial Subtask Decomposition.}
Figure~\ref{fig:subtask_semantics} illustrates SPR's task decomposition capabilities. (d) shows accurate decomposition of basic pick-and-place with semantic descriptions and spatial coordinates. (c) demonstrates understanding of relative positioning constraints. (b) validates robustness under distribution shift: SPR maintains accurate subtask decomposition when object layouts deviate from training configurations. (a) showcases generalization to complex sequential tasks like door closing, correctly identifying distinct manipulation phases and spatial goals.

\noindent\textbf{Error Recovery through Rewind.}
Figure~\ref{fig:robust_drawer} (c) demonstrates handling execution failures that create unexpected state transitions. When the mug grasp fails but the arm continues to the plate position, standard VLA models struggle with this severe distribution shift—the robot reaches the target without the required object. SPR's progress monitoring detects the anomaly (subtask count unchanged despite reaching the goal) and triggers Rewind, returning the arm to the initial position to successfully re-grasp.

\noindent\textbf{Robustness to Unexpected Perturbations.}
Figure~\ref{fig:robust_drawer} (a) highlights dynamic replanning under environmental perturbations. When the bowl inadvertently closes the drawer and falls to a new location, SPR's spatial awareness enables rapid adaptation. The model relocalizes the bowl, updates spatial subtask coordinates, and completes the task, demonstrating spatial awareness that extends beyond in, distribution scenarios.

\noindent\textbf{Handling Challenging Initial Configurations.}
Figure~\ref{fig:robust_drawer} (b) shows robustness to suboptimal initial poses, a key challenge in LIBERO-Plus Robot. The arm begins with poor alignment to the book, and the first grasp fails. Unlike baselines that persist with failures or knock over objects, SPR's Rewind mechanism detects the failure and returns to the initial position, enabling a successful approach from a better angle. This autonomous recovery is valuable for deployment scenarios with variable initial conditions.

\subsection{Failure Analysis}

Figure~\ref{fig:failure_case} illustrates representative failure modes of our approach and their underlying causes. Despite the robust performance of our SPR framework, we identify three primary failure patterns that highlight remaining challenges in vision-language-action models.

\textbf{Precision limitation from discrete action tokens.} Our model outputs actions as discrete token sequences, which inherently limits the precision of continuous control commands. This discretization, while beneficial for leveraging language model architectures, introduces quantization errors that become particularly problematic for tasks requiring fine-grained manipulation. As shown in Figure~\ref{fig:failure_case}(a), when tasked with placing a mug precisely at the center of a plate, the model instead places it near the edge. This occurs because the discrete action space cannot adequately capture the subtle control nuances needed for millimeter-level precision. Such failures are especially prevalent in tasks involving careful object placement, alignment, or insertion operations where small deviations lead to task failure.

\textbf{Rewind ineffectiveness when physically stuck.} While our progress-aware rewind mechanism successfully detects anomalies through subtask count monitoring, it fundamentally relies on the robot's ability to execute actions that change its state. When the robot becomes physically constrained or stuck during execution, the rewind mechanism fails because the physical obstruction prevents meaningful state transitions. Figure~\ref{fig:failure_case}(b) demonstrates this limitation: when attempting to place a mug in a microwave, the mug becomes lodged at the microwave's edge. Despite the model issuing rewind commands, the physical constraint prevents the robot from moving, leaving the subtask count unchanged and the anomaly detection unable to trigger proper recovery. This failure mode reveals a critical gap between progress awareness and physical state awareness—our system can detect logical inconsistencies but cannot directly sense physical impediments.

\textbf{Persistent failure despite successful rewind.} Perhaps most concerning are cases where our model correctly identifies execution errors, successfully triggers and executes the rewind procedure, yet remains unable to complete the task. Figure~\ref{fig:failure_case}(c) illustrates this complex failure pattern during a grasp bbq sauce task. Initially, the model encounters spatial misalignment between the gripper and target object, correctly triggering a rewind. However, after rewinding, the model fails to correct the misalignment. In subsequent attempts, even after returning to the initial position, the model exhibits hallucination behavior with significant positional deviations from the target object. Most troublingly, even in instances where the predicted 2D trajectory and subtask waypoints accurately indicate the correct target location, the generated actions fail to align with these planned waypoints, resulting in the gripper moving to different positions than intended. This reveals a fundamental disconnect between the model's spatial planning capabilities and its action execution—while it can correctly perceive and plan where to go, the actions it generates do not faithfully follow these plans. Such failures indicate that spatial awareness and trajectory planning alone are insufficient without ensuring consistency between planned waypoints and executed actions.

\section{Conclusion}

We introduce \textbf{S}ee, \textbf{P}lan, \textbf{R}ewind (SPR), a vision-language-action framework for robust manipulation via spatial subtask decomposition and autonomous recovery.  
Extensive experiments validate our approach: SPR outperforms MolmoAct by 5\% on LIBERO. On the challenging LIBERO-Plus benchmark, it achieves state-of-the-art robustness, exhibiting the smallest performance degradation (\textit{–18.8\%} in average) across unseen instructions and initial states perturbations, underscores its superior generalization.
Future work will be extending our method to diverse simulations, and real-world scenarios while handling noisy or suboptimal training demonstrations.

\bibliographystyle{IEEEtran}
\bibliography{main}  

\clearpage
\end{document}